\documentclass[11pt]{article}

\usepackage{arxiv}

\usepackage[utf8]{inputenc}
\usepackage{blindtext}
\usepackage{booktabs}
\usepackage{tikz}
\usepackage{xurl}
\usepackage{algorithm}
\usepackage{algorithmicx}
\usepackage{algpseudocode}
\usepackage{subcaption}
\usepackage{natbib}
\usepackage[export]{adjustbox}
\usepackage{hyperref} 
\usepackage{url}  
\usepackage{amsfonts}
\usepackage{nicefrac}
\usepackage{microtype}
\usepackage{cleveref}
\usepackage{lipsum}
\usepackage{graphicx}
\usepackage{doi}

\usetikzlibrary{positioning}
\usetikzlibrary{calc}
\usetikzlibrary{bending}
\usetikzlibrary{arrows.meta}

\title{High-Quality Tabular Data Generation using Post-Selected VAE}

\author{
  \href{https://orcid.org/0009-0000-2697-8486}
  {Volodymyr Shulakov} \\
	Department of Computer Science and System Analysis\\
	Cherkasy State Technological University\\
	Shevchenka Blvd, 460, Cherkasy, Cherkasy Oblast, 18000, Ukraine \\
	\texttt{v.v.shulakov.fitis19@chdtu.edu.ua} \\
}

\begin{document}
\maketitle

\begin{abstract}
  Synthetic tabular data is becoming a necessity as concerns about data privacy intensify in the world.
  Tabular data can be useful for testing various systems, simulating real data,
  analyzing the data itself or building predictive models. Unfortunately, such data may not be
  available due to confidentiality issues. Previous techniques, such as TVAE (\citealp{tvae}) or OCTGAN (\citealp{octgan}),
  are either unable to handle particularly complex datasets, or are complex in themselves,
  resulting in inferior run time performance.
  This paper introduces PSVAE, a new simple model that is capable of
  producing high-quality synthetic data in less run time. PSVAE
  incorporates two key ideas: loss optimization and post-selection. Along with these ideas,
  the proposed model compensates for underrepresented categories and uses a modern activation function,
  Mish (\citealp{misra2019mish}).
\end{abstract}

\keywords{
  generative models \and variational autoencoder \and synthetic data \and multivariate sampling \and tabular synthesis
}

\section{Introduction}

The importance of maintaining privacy of user data continues to grow today,
and real data, including tabular data that can be used in applied data modelling
(such as building a predictive model), socio-economic and physical processes, is confidential.
Usage of such data in information systems is therefore becoming increasingly difficult.
Techniques based on a combination of statistical methods and neural networks
are being incorporated into modern data generation systems.
As a result, implementation of such generative algorithms for publishing data
that is completely different numerically but similar to the original statistical
characteristics, is becoming increasingly necessary.

In particular, synthetic data is relevant for a variety of purposes, such as testing and validating machine
learning algorithms, testing integration software (\citealp{taneja}), building simulation models,
researching statistical characteristics of data in various fields, such as healthcare (\citealp{alsmadi}).
The use of synthetic data generation systems makes it possible to obtain data for a research without directly
using private data.

Synthetic data generation is not just an innovation, but a solution for accurate, secure,
and cost-effective data modelling. According to Gartner, synthetic data will replace real data
in artificial intelligence by 2030 (\citealp{gartner}).

The objective of this work is to develop a simple and robust method for generating synthetic data of high quality,
thereby providing a model for integration into various systems for a wide range of needs.
The model presented here, PSVAE, builds upon the variational autoencoder (\citealp{kingmamw}) architecture
and employs a custom loss adjustment algorithm and a bespoke post-selection
mechanism to filter the output of the decoder network. This approach allows for extraction of
the optimal univariate statistics from complex datasets, while maintaining the quality of multivariate relationships.
The algorithm can simulate both discrete and continuous distributions.
It appears that this concept of post-selection has not previously been employed in this manner.

\section{Related Work}

The most basic approach to generating random numerical data according to some distribution
is to try to mimic the distribution by closely matching its univariate distributions.
Doing so neglects the most crucial part, namely the dependencies between variables, which represent
the essence of high-quality synthetic data.

The recent advances in deep generative models have opened up a wide range of possibilities.
Specifically, powerful neural network-based methods have emerged, most notably CTGAN and TVAE (\citealp{tvae})
which can learn better distributions than Bayesian networks such as those proposed by \cite{chowliu} or \cite{privbayes}.
\cite{octgan} proposed OCT-GAN, a generative model based on neural ordinary differential
equations that outperforms TVAE in some cases. The synthesis quality of Invertible Tabular GANs
(\citealp{leejaehoon}) is comparable to that of TVAE. \cite{afonja} introduce MargCTGAN,
an improved version of CTGAN for generating synthetic tabular data, particularly in low-sample regimes.
CTAB-GAN+ (\citealp{ctabgan}) uses Wasserstein loss with gradient penalty for better training convergence
and adds downstream losses to conditional GANs for higher utility synthetic data.

It seems reasonable to posit that the VAE architecture offers an elegant and efficient solution
for developing synthetic data generative models. Consequently, this work is focused
on the implementation of this type of neural network in order to achieve this objective.

\section{Post-Selected VAE-based Data Generator}

PSVAE builds on a VAE and is similar to TVAE, but does not include specific layers for `multi-modal' normalization.
The encoder transforms multiple one-hot encoded categories through
two feature-extracting 256-neuron linear layers.
These features are then passed through another two 256-neuron layers to obtain 128 $\mu$
and 128 $\sigma$ values, which account for one input layer and two output layers.
$\mu$ and $\sigma$ are used to re-parameterize the latent space. No activation is applied after the output layers.
The decoder differs only in that its input dimension is of the size of the latent space (128 float values)
and output size is the same as the encoder's input size.
Like the authors of TVAE, we found that a batch size of 500 performs the best (\citealp{tvae}).
To generate continuous synthetic variables, real data is simply discretized into buckets.
The number of buckets is $\min(\sqrt N, 100)$ where N is the number of records in a dataset.

To improve generalization, in-between hidden layers a modern smooth, continuous, self regularized,
non-monotonic activation function called {\it mish} is used, which provides better empirical results
than Swish, ReLU, and Leaky ReLU (\citealp{misra2019mish}).
It is incorporated into most ML frameworks, including Pytorch. In PSVAE it is used instead of ReLU, in contrast to TVAE.
Mish function is defined as follows.
\[ f(x) = x \tanh(\ln(1+e^x)). \]

As with the regular VAE implementation, reconstruction loss (categorical cross entropy)
and KL-divergence of $\mu$ and $\sigma$ are used in combination.
However, in the case of TVAE, a factor of 2 is applied to the reconstruction loss, which
has the effect of impairing the model's performance on certain datasets.
In order to achieve balance between these two losses, a simple algorithm has been
devised. This balancing takes inspiration from $\beta$-VAE
(\citealp{higgins2017beta}), where our approach does not just tune $\beta$, but optimizes it
automatically in every training session (Algorithm \ref{alg:loss}). It also improves the performance of the
post-selection mechanism.
\[ Loss = L_{RE} + \beta L_{KL}, \]
where $L_{RE}$ is the reconstruction loss and $L_{KL}$ is the regularization loss.

To address the issue of imbalanced datasets like {\it credit}, the cross-entropy loss
is weighted by the inverse frequency of the category occurrence ($\omega_i$)
\[ L(x,y) = - \sum_{c=1}^C { \omega_i \log \frac{ e^{x_c} }{ \sum_{i=1}^{C} { x_i }} * y_c}. \]

Experimental results indicate that this approach yields superior results
in terms of synthetic data quality compared to the TVAE architecture.

\begin{algorithm}[]
  \caption{PSVAE loss adjustment algorithm}
  \label{alg:loss}

  \begin{algorithmic}
    \State $\theta \gets$ network parameters
    \State $\beta \gets 1$

    \For{each epoch}

    \State $L_{KL} \gets 0$
    \State $L_{R} \gets 0$

    \For{each batch}
    \State $l_{kl}, l_{r} \gets \Call{model}$
    \State $L \gets L_{KL} + \beta L_{RE}$
    \State $\theta \gets \Call{optimize}{\theta, L}$
    \State $L_{KL} \gets L_{KL} + l_{kl}$
    \State $L_{R} \gets L_{R} + l_{r}$
    \EndFor

    \State $s \gets L_{KL} + L_{RE}$
    \State $d \gets 1/3 * s$

    \If{$ L_{KL} > d $}
    \State $\beta \gets \beta * 1.04$
    \Else
    \State $\beta \gets \beta / 1.04$
    \EndIf

    \EndFor
  \end{algorithmic}
\end{algorithm}

\begin{figure}[]
  \centering
  \begin{tikzpicture}[
      font = \fontsize{10.5pt}{10.5pt}\selectfont,
      node distance = 10mm,
      process/.style = {rectangle, draw, rounded corners, text width=9em, minimum height=3em, align=center, fill=cyan!10},
      data/.style = {rectangle, draw, rounded corners, text width=9em, minimum height=3em, align=center, fill=green!20},
      arrow/.style = {thick, ->, >=stealth}
    ]
    % Nodes
    \node [process] (latents) {Sampling random latent variables};
    \node [process, right=of latents] (vae) {Decoding};
    \node [process, below=of vae] (post) {Post-Selection};
    \node [data, right=16em of $(vae.south)!0.5!(post.south)$, sharp corners, anchor=south]
    (out) {Synthetic data batch};

    % Arrows
    \draw [arrow] (latents) -- (vae);
    \draw (post) edge[arrow, bend left] node[fill=white] {N cycles} (latents);
    \draw (vae) edge[arrow] (post);

    \draw[arrow, style={rounded corners}] (vae) -| node[above, pos=0.28, fill=white] {Initialization} (out);
    \draw[arrow, style={rounded corners}] (post) -| node[below, pos=0.28, fill=white] {Improvement} (out);

  \end{tikzpicture}
  \caption{Illustration of the synthetic data generation workflow of PSVAE}
  \label{fig:genflow}
\end{figure}
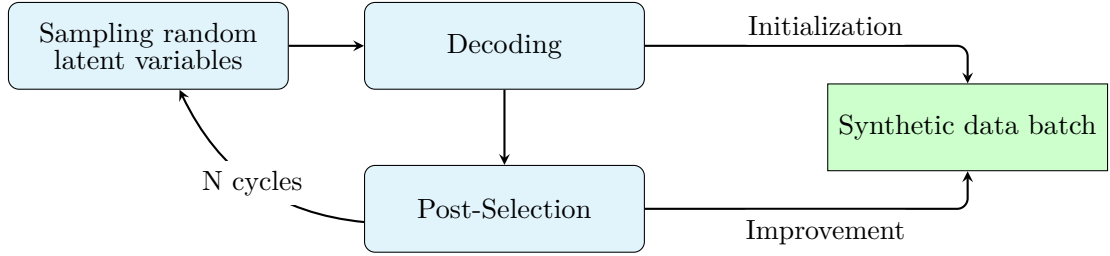

One of the most important parts of PSVAE is post-selection (Figure \ref{fig:genflow}). This mechanism plays the role
of enhancing VAE output. To achieve this, an initial batch of synthetic data is sampled from the decoder and
is refined by replacing samples with new ones from subsequent generated batches, provided that the new samples
exert a positive influence on the distribution of samples (Algorithm \ref{alg:postsel}).

\begin{algorithm}[h]
  \caption{PSVAE post-selection algorithm}
  \label{alg:postsel}

  \begin{algorithmic}
    \State {$V \gets$ normally-distributed random 128 floats}
    \State $S \gets$ \Call{decoder}{V}
    \State $P \gets$ \Call{pdf}{S}

    \For{each selection cycle}

    \State {$V \gets$ normally-distributed random 128 floats}
    \State $R \gets$ \Call{decoder}{V}

    \For{$i=1$ to $n$} \Comment {for each sample in the batch}

    \State $c \gets S_i$
    \State $r \gets R_i$

    \If{ \Call{influence}{$P$, $c$, $r$} $ > 0$ } \Comment {check if it improves PDF similarity}
    \State $S_i \gets r$
    \State $P \gets$ \Call{newpdf}{$P$, $c$, $r$}
    \EndIf

    \EndFor

    \EndFor
  \end{algorithmic}
\end{algorithm}

\section{Experiments}

TVAE, OCT-GAN and CTAB-GAN+ are selected to assess their performance
in comparison to PSVAE. Interestingly, despite the post-selection step utilizing only univariate similarity,
PSVAE is able to successfully reproduce statistical intricacies of the original data,
thereby confirming that the quality of learned multivariate dependencies remains intact.

In the course of the experiments, L1-distance, F1 classification score
and Pearson correlation coefficients have been employed. Three datasets have been selected for analysis
(see Table \ref{tab:datasets}).

\begin{itemize}
  \item[--] Brain Stroke. URL: \url{https://www.kaggle.com/datasets/jillanisofttech/brain-stroke-dataset}
  \item[--] Diabetes Health Indicators. URL: \url{https://www.kaggle.com/datasets/alexteboul/diabetes-health-indicators-dataset}
  \item[--] Credit Card Fraud Detection. Has 31 columns, where the last one (Class) takes
        value 1 in case of fraud and 0 otherwise. There are only 0.1\% 1s out of 284k 0s.
        Thus, it is a challenging dataset to reproduce statistically.
        URL: \url{https://www.kaggle.com/datasets/mlg-ulb/creditcardfraud}
\end{itemize}

\begin{table}[h]
  \centering
  \caption{datasets used in experiments}
  \begin{tabular}{llll}
    \toprule

    Name         & \# records & \# variables & Identity F1 \\
    \midrule
    Brain Stroke & 4981       & 11           & 0.57        \\
    Diabetes     & 253680     & 22           & 0.6         \\
    Credit       & 284807     & 31           & 0.9         \\

    \bottomrule
  \end{tabular}
  \label{tab:datasets}
\end{table}

Table \ref{tab:eval} presents evaluations of all datasets of the models that were
selected for analysis. $t_e$ denotes the training run time of an epoch, in seconds.
In order to obtain an L1 metric, all L1-distances across all distributions of columns in a dataset are averaged.
A classifier network is trained on synthetic data and subsequently tested against real data from the
selected datasets to calculate a mean F1-score of all prediction labels. Table \ref{tab:datasets} also
presents Identity-F1 metrics calculated on 20\% of the data of each dataset, with the remaining 80\%
used to train the classifier networks. The sum of all absolute Pearson correlation coefficient differences of
each pair of column data distributions is denoted by $\rho$. Categorical values in the form of alphanumeric labels
are converted to unsigned sequential integers.

PSVAE is trained for 100 epochs for each dataset, with 10 post-selection cycles performed
for each synthetic data batch generation. Adam optimizer is used with a learning rate of $10^{-3}$.
A single batch consists of 500 randomly selected records. During experimentation, the performance
of OCT-GAN and CTAB-GAN was observed to not increase significantly
after training for 50 and 100 epochs, respectively.

\begin{table}[b]
  \centering
  \caption{Evaluation of different synthetic data generation models}
  \begin{tabular}{lccccccccccl}
    \toprule
              & \multicolumn{1}{c}{}
              & \multicolumn{3}{c}{Brain stroke}
              & \multicolumn{3}{c}{Diabetes}
              & \multicolumn{3}{c}{Credit}
    \\ \cmidrule(l){3-5} \cmidrule(l){6-8} \cmidrule(l){9-11}

    Model     & $t_e$                            & L1        & F1        & $\rho$    & L1        & F1       & $\rho$    & L1        & F1        & $\rho$    \\
    \midrule
    OCT-GAN   & 30                               & 0.19      & 0.5       & 5.12      & 0.14      & 0.38     & 10.6      & 0.08      & \bf{0.84} & 25.1      \\
    CTAB-GAN+ & 5                                & 0.08      & 0.51      & 3.46      & 0.18      & 0.41     & \bf{6.07} & 0.07      & \bf{0.82} & 23.5      \\
    TVAE      & \bf{1}                           & 0.14      & 0.49      & 5.54      & 0.11      & \bf{0.6} & 67.6      & 0.08      & 0.51      & 32.4      \\
    \midrule
    PSVAE     & \bf{2}                           & \bf{0.01} & \bf{0.55} & \bf{1.76} & \bf{0.01} & 0.45     & 12.1      & \bf{0.01} & \bf{0.83} & \bf{15.8} \\

    \bottomrule
  \end{tabular}
  \label{tab:eval}
\end{table}

A modern GPU (RTX 4060) was used to train all the models.
OCT-GAN and CTAB-GAN+ took considerable time to achieve
performance comparable to that of PSVAE. When training on the {\it diabetes} dataset,
a single epoch took 2 seconds for PSVAE, whereas CTAB-GAN+ took 5 seconds per epoch and
OCT-GAN performed a single epoch in 30 seconds. This slowdown may be due to high complexity
of the underlying neural architectures.

\begin{figure}[t]
  \adjustbox{minipage=1.3em,valign=t}{\subcaption{}\label{fig:braincorr}}
  \begin{subfigure}[]{\dimexpr0.95\linewidth-0em}
    \centering
    \includegraphics[width=1\linewidth,valign=t]{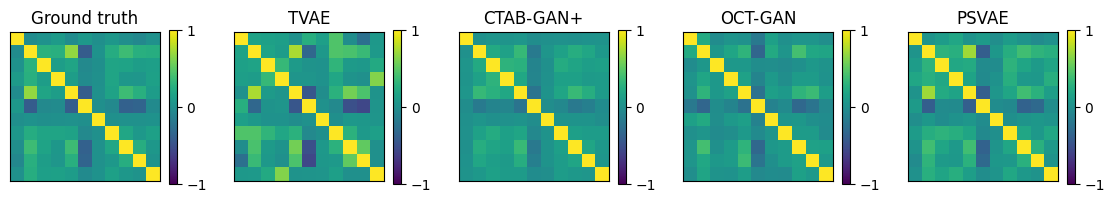}
  \end{subfigure}

  \adjustbox{minipage=1.3em,valign=t}{\subcaption{}\label{fig:diabetescorr}}
  \begin{subfigure}[]{\dimexpr0.95\linewidth-0em}
    \centering
    \includegraphics[width=1\linewidth,valign=t]{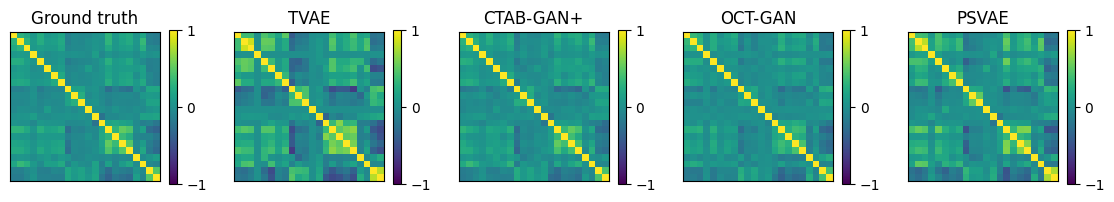}
  \end{subfigure}

  \adjustbox{minipage=1.3em,valign=t}{\subcaption{}\label{fig:creditcorr}}
  \begin{subfigure}[]{\dimexpr0.95\linewidth-0em}
    \centering
    \includegraphics[width=1\linewidth,valign=t]{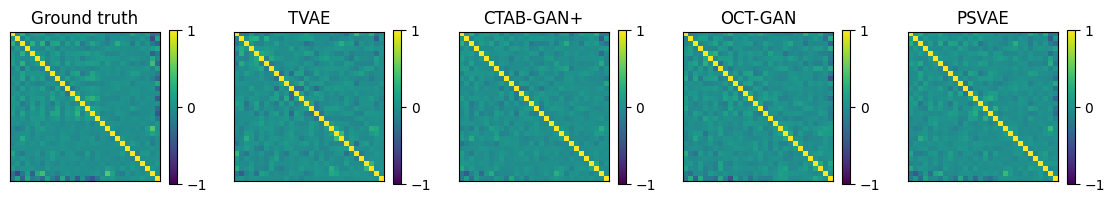}
  \end{subfigure}

  \caption{Correlations between variables of (a) {\it brain stroke} dataset,
    (b) {\it diabetes} dataset, (c) {\it credit} dataset. It can be seen in (c) that TVAE does not have
    proper correlations in the last column.}
  \label{fig:corr}
\end{figure}

As illustrated in Figure \ref{fig:corr},
PSVAE is capable of capturing all the intricate details of the original datasets.
The previous models are unable to identify much of the correlations present in the last column
of the {\it credit} dataset, likely due to imbalanced data distribution.
Interestingly, TVAE has better F1 performance, likely due to data distillation by
approximating continuous variables with Gaussian mixtures.

The results of ablation study are as follows.
\begin{itemize}
  \item[--] {Replacing Mish activation with ReLU results in higher losses and thus slower training time.}
  \item[--] {Removal of either categorical loss weighting or loss adjustment algorithm precludes the model from
        learning intricate correlations ($\rho$), such as those observed in the {\it credit} dataset.}
  \item[--] {Removing the post-selection deteriorates the L1 accuracy down to the level of TVAE, requiring
        a larger network and a longer training time to address underfitting.}
\end{itemize}

\section{Conclusion}

The study presents a new approach to generating high-quality synthetic tabular data called PSVAE that
augments a VAE with improved loss computation algorithm, data balancing, modern activation function {\it mish}
and a post-selection mechanism.

Ultimately, PSVAE offers superior performance on L1-distance,
comparable F1-scores, and in some cases better correlation synthesis
of datasets of varying complexity, compared to previous solutions.
TVAE tends to poorly reproduce complex data correlations,
although OCT-GAN manages to achieve a good $\rho$-distance on the {\it diabetes} dataset.
Nevertheless, previous methods that produce competitive results
have inferior run times in comparison to PSVAE.

In future work, vector-quantization may be used (\citealp{van2017neural}) to investigate its
capabilities within VAE in generating synthetic data of even better quality.

\vskip 0.2in
\bibliography{main}

\end{document}